\documentclass[10pt,twocolumn,letterpaper]{article}

\usepackage[pagenumbers]{cvpr} 
\usepackage{times}
\usepackage{epsfig}
\usepackage{graphicx}
\usepackage{amsmath}
\usepackage{amssymb}
\usepackage{caption}
\usepackage{booktabs, siunitx}
\usepackage{multirow, booktabs}
\usepackage{siunitx}
\usepackage{makecell} 

\usepackage{lipsum}
\usepackage{fontawesome}
\usepackage{mathtools}

\newcommand{\defeq}{\coloneqq}

\newcommand{\E}{\mathbb{E}}

\newcommand{\Eb}[2]{\E_{#1}\!\left[#2\right]}

\newcommand{\bI}{\mathbf{I}}

\newcommand{\bzero}{\mathbf{0}}

\newcommand{\bx}{\mathbf{x}}

\newcommand{\bepsilon}{{\boldsymbol{\epsilon}}}

\DeclareMathOperator*{\argmin}{arg\,min}

\usepackage{marvosym, ifsym}
\usepackage{ulem}

\usepackage[dvipsnames]{xcolor}

\definecolor{cvprblue}{rgb}{0.21,0.49,0.74}
\usepackage[pagebackref,breaklinks,colorlinks,citecolor=cvprblue]{hyperref}

\def\paperID{3560} 
\def\confName{CVPR}
\def\confYear{2024}

\begin{document}
\title{CatVersion: Concatenating Embeddings for Diffusion-Based \\Text-to-Image Personalization}





\author{
  Ruoyu Zhao$^{1}$ \quad Mingrui Zhu$^{1}$ \quad Shiyin Dong$^{1}$  \quad Nannan Wang$^{1}$\textsuperscript{\faEnvelopeO}  \quad Xinbo Gao$^{2}$\\
  $^1$ Xidian University \\
  $^2$ Chongqing University of Posts and Telecommunications \\
  \small{\url{https://royzhao926.github.io/CatVersion-page/}}
}


\twocolumn[{%
            \renewcommand\twocolumn[1][]{#1}%
            \maketitle
            \vspace{-3em}
            \begin{center}
                \centering
                \includegraphics[width=1\textwidth]{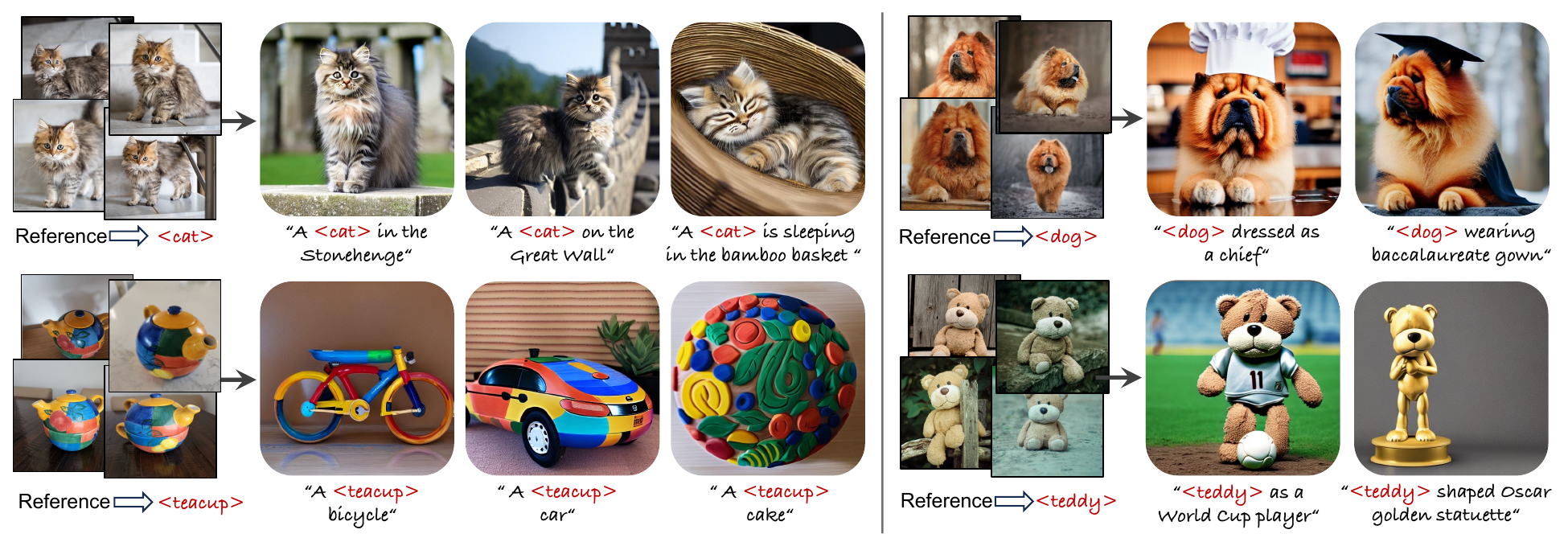}
                \vspace{-15pt}
                \captionof{figure} {\textbf{CatVersion} allows users to learn the personalized concept through a handful of examples and then utilize text prompts to generate images that embody the personalized concept. In contrast to existing approaches, CatVersion \textbf{concatenates} embeddings on the \textbf{feature-dense space} of the text encoder in the diffusion model to learn the gap between the personalized concept and its base class, aiming to maximize the preservation of prior knowledge in diffusion models while restoring the personalized concepts. }
                \label{teaser}
            \end{center}%
        }]
\maketitle

\newcommand\blfootnote[1]{%
\begingroup
\renewcommand\thefootnote{}\footnote{#1}%
\addtocounter{footnote}{-1}%
\endgroup
}
\blfootnote{\faEnvelopeO~ Corresponding author.}

\begin{abstract} 

We propose \textbf{CatVersion}, an inversion-based method that learns the personalized concept through a handful of examples. Subsequently, users can utilize text prompts to generate images that embody the personalized concept, thereby achieving text-to-image personalization.
In contrast to existing approaches that emphasize word embedding learning or parameter fine-tuning for the diffusion model, which potentially causes concept dilution or overfitting, our method \textbf{concatenates} embeddings on the \textbf{feature-dense space} of the text encoder in the diffusion model to learn the gap between the personalized concept and its base class, aiming to maximize the preservation of prior knowledge in diffusion models while restoring the personalized concepts.
To this end, we first dissect the text encoder's integration in the image generation process to identify the feature-dense space of the encoder. Afterward, we concatenate embeddings on the Keys and Values in this space to learn the gap between the personalized concept and its base class. In this way, the concatenated embeddings ultimately manifest as a residual on the original attention output.
To more accurately and unbiasedly quantify the results of personalized image generation, we improve the CLIP image alignment score based on masks. Qualitatively and quantitatively, CatVersion helps to restore personalization concepts more faithfully and enables more robust editing.

\end{abstract}

\section{Introduction}
\label{sec:intro}

Recently, text-guided diffusion models \cite{nichol2021glide,ramesh2022hierarchical,saharia2022photorealistic,rombach2022high} have garnered significant attention due to their remarkable high-fidelity image synthesis capabilities. These models utilize natural language descriptions to synthesize high-quality images aligned with these texts. However, they still encounter challenges when dealing with personalized concepts that are difficult to describe accurately.

\begin{figure}[t]
  \centering
   \includegraphics[width=0.48\textwidth]{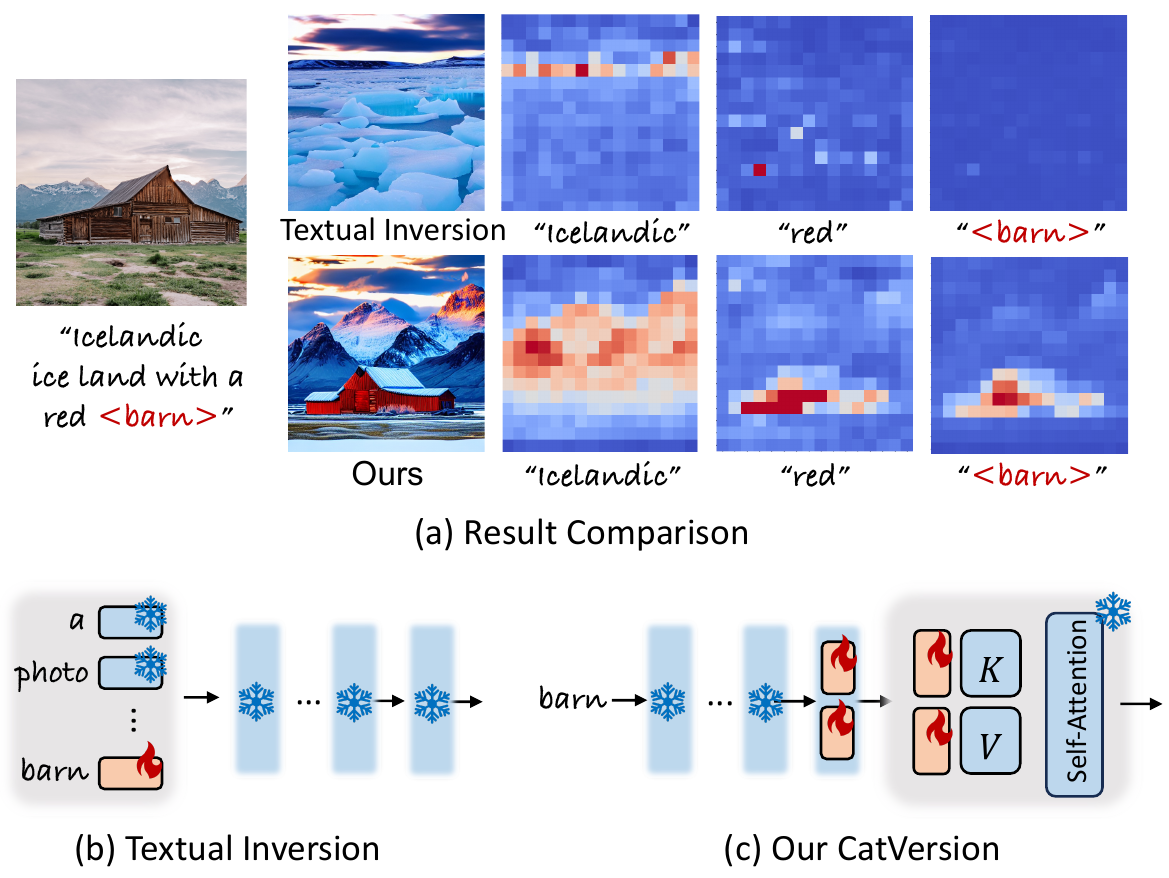}
   \caption{\textbf{CatVersion versus Textual Inversion}. We showcase the results of Textual Inversion \cite{gal2022image} with our CatVersion. As shown in (a), Textual Inversion fails to capture the personalized concept, while CatVersion accurately restores it. We contrast the distinctions in the inversion spaces of the two methods in (b) and (c), underscoring the advantages of inversion in feature-dense space.
   }
   \label{fig:attention_map}
\end{figure}

Text-to-image (T2I) personalization \cite{gal2022image,ruiz2023dreambooth,kumari2023multi} offers unprecedented opportunities for users to describe personalized concepts. With a handful of examples representing one concept, users can employ free-text descriptions to synthesize images depicting the concept. Based on the extensive prior knowledge derived from text-guided diffusion models, recent T2I personalization schemes \cite{gal2022image,huang2023reversion} invert the concept by optimizing word embeddings corresponding to a pseudo-word, and then combine the pseudo-word with free text to create a new scene of the concept. However, we observe that during pseudo-word guided synthesis, especially when combining with free-text, the personalized concept is prone to becoming diluted or lost, as shown in Figure \ref{fig:attention_map}.

This is because optimizing word embeddings with few examples to represent the personalized concept accurately is highly challenging. Voynov et al. \cite{voynov2023p+} and Zhang et al. \cite{zhang2023prospect} optimize multiple word embedding instances for the multi-scale blocks of the U-Net backbone or different time steps of the denoising process, providing more precise control. However, the challenge of optimizing word embeddings has not yet been alleviated. 
To avoid this challenge, Ruiz et al. \cite{ruiz2023dreambooth} and Kumari et al. \cite{kumari2023multi} resort to fine-tuning all or part of the parameters of the network for aligning rare word embeddings with the target concept. However, this will undermine the prior knowledge of the diffusion model and lead to overfitting in personalized concept generation.

We introduce \textbf{CatVersion}, a prompt inversion method that \textbf{concatenates} embeddings into a \textbf{highly integrated feature space}, which helps to restore the personalized concept more faithfully and enabling more robust editing.
In contrast to directly optimizing word embeddings corresponding to personalized concepts, CatVersion learns the gap between the concept and its base class by concatenating learnable embeddings to the Keys and Values in the highly integrated feature space of the CLIP text encoder, facilitating a more effective inversion process. 

Specifically, we show that various attention layers of the CLIP text encoder emphasize different aspects in the text-guided diffusion generation process. Shallow layers mainly focus on the construction of subject information, while as the layers deepen, more other complex and abstract information are gradually integrated. Based on this integration, we locate the highly integrated feature space within the last few attention layers of the CLIP text encoder where learning personalized concepts is more effective. Then, we concatenate personalized embeddings to the Keys and Values in these layers and optimize them. Unlike word embeddings that learn personalized concepts directly, these personalized embeddings are ultimately represented as a residual on the original attention output to learn the gap between the personalized concept and its base class. We refer to the personalized embeddings learned in this way as ``\textbf{residual embeddings}''. Like word embeddings, residual embeddings are plug-and-play and can be combined with free text to apply personalized concepts to different scenarios.

In T2I personalization evaluation, it is crucial to carry out objective quantitative research. We analyze the CLIP image alignment score and find that it does not adapt well to personalized generation tasks. So we improve it to make an accurate and unbiased evaluation.

In summary, our contributions are threefold:
\begin{itemize}
    \item We analyze the integration of the CLIP text encoder in T2I diffusion models and introduce a tightly integrated feature space that facilitates the concept inversion.
    \item We propose CatVersion, a straightforward yet effective inversion method. It concatenates embeddings to the Keys and Values within the tightly integrated feature space of the text encoder to learn the gap between the concept and its base class as residuals.
    \item To quantify the results more accurately and unbiased, we adjust the CLIP image alignment score to make it more rational. Extensive experiments, both qualitatively and quantitatively, demonstrate the effectiveness of our method in faithfully restoring the target concepts and enabling more robust editing.
\end{itemize}

\begin{figure*}[t]
  \centering
   \includegraphics[width=1\textwidth]{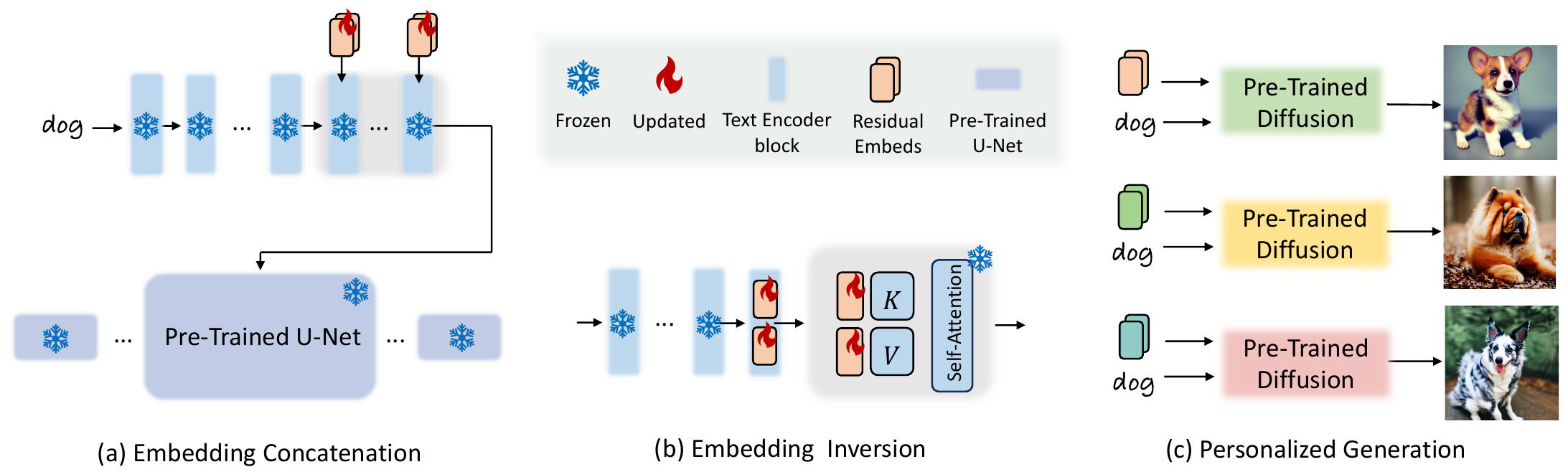}
   \caption{\textbf{Overall Pipeline of CatVersion}. Firstly, we identify the feature-dense layers in the CLIP text encoder. Then, we concatenate the residual embeddings with Keys and Values. In the optimization process, we use the base class word (e.g. dog) of the personalized concept as text input and optimize these residual embeddings utilizing a handful of images depicting one personalized concept. During inference, residual embeddings of CatVersion can be deleted and replaced to achieve different personalized needs.
   }
   \label{fig:framework}
\end{figure*}

\section{Related Work}
\subsection{Text-to-Image Diffusion Models}
Text-to-image generation has been extensively studied in the past few years. Prior efforts mainly focus on GAN-based \cite{zhang2017stackgan,zhang2018stackgan++,qiao2019mirrorgan,xu2018attngan,yin2019semantics,zhu2019dm,tao2022df} and autoregressive network-based \cite{ramesh2021zero,ding2021cogview,lin2021m6,gafni2022make} methods. The former utilizes multimodal visual-language learning such as CLIP \cite{radford2021learning} to achieve semantic alignment between text description and generated images, demonstrating satisfactory results in text-guided image generation. However, its training process is relatively unstable and prone to mode collapse. Based on the transformer architecture, the latter adopts a tokenizer like VQ-VAE \cite{van2017neural}, converting the input text description into continuous vector representations and then generating high-fidelity images based on these representations as conditions. Although it exhibits excellent performance in text-to-image generation quality, its training process requires significant computing resources and memory usage.

In recent years, diffusion models \cite{ho2020denoising,nichol2021improved,song2020denoising,song2020score,dhariwal2021diffusion}, as score-based generative models, have garnered significant acclaim owing to their stunning image-generation capabilities. Therefore, they have quickly become a new frontier in image synthesis.
Text-guided diffusion models use text as an extra condition to guide image generation. 
By training massive text-image data pairs, it can associate text embeddings with the feature of the image, thereby guiding the image generation process.
Recent works such as GLIDE \cite{nichol2021glide}, DALL-E 2 \cite{ramesh2022hierarchical}, Imagen \cite{saharia2022photorealistic}, and Stable Diffusion \cite{rombach2022high} have demonstrated the remarkable performance of text-guided diffusion models in generating diverse and high-fidelity images. We extensively utilize the prior knowledge of these models for T2I personalization.

\subsection{Diffusion-based T2I personalization}
The personalized concept is often abstract and difficult to express accurately using text descriptions. When using the T2I diffusion models to synthesize the image with these concept, obstacles are often encountered. T2I personalization aims to learn abstract concepts from a handful of sample images and apply these concepts to new scenarios.

Gal et al. \cite{gal2022image} learn the personalized target by optimizing textual embeddings corresponding to the pseudo-word. The pseudo-word can be combined with free text to guide the personalized generation during inference. 
Ruiz et al. \cite{ruiz2023dreambooth} use a class noun combined with a unique identifier to represent the target concept. The unique identifier is represented by rare tokens. To invert the target concept into the rare token, they fine-tune the diffusion model with a prior-preservation loss. 
Based on this method, Kumari et al. \cite{kumari2023multi} only update the weight of cross attention to invert the concept onto the rare token. In addition, they select a regularization set to prevent overfitting. 
Tewel et al. \cite{tewel2023key} propose Perfusion, using a gated Rank-one Model Editing \cite{meng2022locating} to the weights of the Key and Value projection matrices in cross-attention of the U-Net and reducing concepts leakage beyond their scope, making it easier to combine multiple concepts.
Gal et al. \cite{gal2023encoder} and Wei et al. \cite{wei2023elite} focus on using extensive data to build an encoder for concept inversion.
Huang et al. \cite{wen2023hard} focus on learning the relation between objects through embedding optimization and the relation-steering contrastive learning scheme. 
Wen et al. \cite{wen2023hard} invert hard prompts by projecting learned embeddings onto adjacent interpretable word embeddings, providing a new solution for image captioning.  
Han et al. \cite{han2023svdiff} use singular value decomposition to fine-tune the singular value matrix of the diffusion model network, reducing the number of parameters needed for semantically aligning rare tokens with target concepts. Voynov et al. \cite{voynov2023p+} optimize multiple word embeddings for modules with different feature dimensions of the denoising network, while Zhang et al. \cite{zhang2023prospect} optimize multiple word embeddings for various time steps of denoising. Both methods provide finer control over the generated image, allowing for more precise and accurate output.

These methods either learn word embeddings at the text encoder's input to represent personalized concepts, or fine-tune all or part of the parameters of the diffusion model to align the rare token with the personalized concept. However, the challenge that word embeddings are difficult to optimize is not addressed.

\begin{figure*}[t]
  \centering
  \vspace{-10pt}
   \includegraphics[width=1\linewidth]{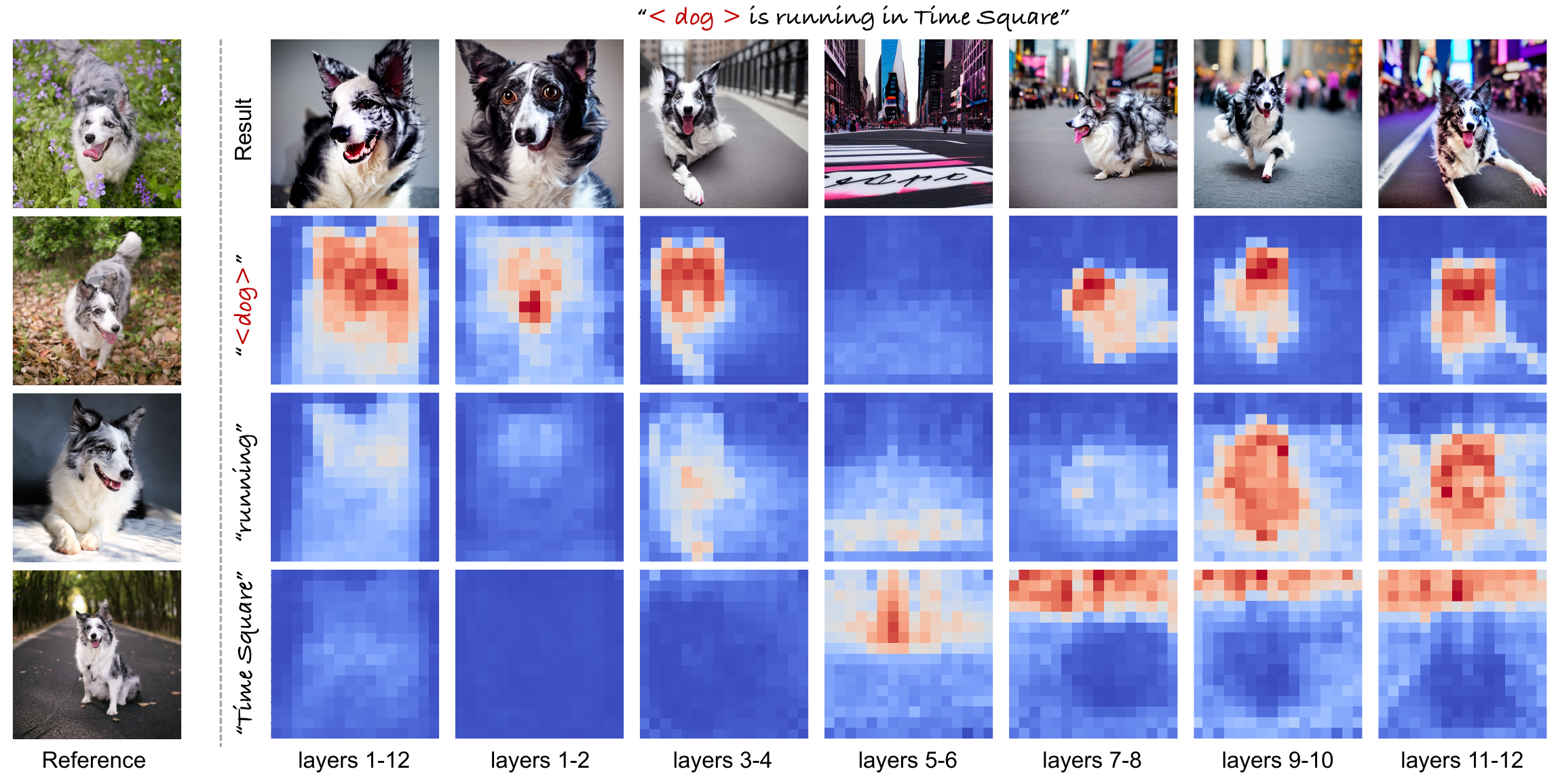}
   \vspace{-15pt}
   \caption{\textbf{Visualizing Inversion across Multiple Layers}. We concatenate embeddings and optimize them in each of the two self-attention layers in the CLIP text encoder. Then, we use these embeddings in combination with free text to create new scenarios for personalized concepts. The results indicate that the self-attention layers of different depths focus on integrating different information. Moreover, the focus of information integration has also shifted from concreteness to abstraction.}
   \vspace{-5pt}
   \label{fig:ablation_dog}
\end{figure*}
\section{Preliminaries}
\subsection{Latent Diffusion Models}
Diffusion models \cite{ho2020denoising,song2020score,nichol2021improved,song2020denoising} are a class of generative models. They convert the input Gaussian noise into sample images that match the target distribution through iterative denoising. Diffusion models allow for conditional guided generation based on category labels \cite{dhariwal2021diffusion}, text \cite{nichol2021glide,ramesh2022hierarchical}, or images \cite{saharia2022palette}. 
The simplified optimization objective for training diffusion models is as follows:
\begin{gather}
    L_\mathrm{DM}(\theta) \defeq \Eb{t, \bx_0, \bepsilon}{ { \left\| \bepsilon - \bepsilon_\theta(\bx_t, t) \right\|^2}}, \label{eq:dm_loss}
\end{gather}
Here, $\bx_t$ denotes the noisy image at time step $t$. It constructs by adding noise $\bepsilon \sim \mathcal{N}(\bzero, \bI)$ to the nature image $\bx_0$. $\bepsilon_\theta(\cdot)$ denotes the noise predicted by the neural network.

Latent diffusion model (LDM) \cite{rombach2022high} is a text-guided diffusion model. 
It uses an encoder $\mathcal{E}(\cdot)$ to map images to latent space and perform an iterative denoising process. Afterward, the predicted images are mapped back to pixel space through the decoder $\mathcal{D}(\cdot)$. 
The simplified optimization objective of LDM is as follows:
\begin{gather}
    L_\mathrm{LDM}(\theta) \defeq \Eb{t, \bx_0, \bepsilon}{ { \left\| \bepsilon - \bepsilon_\theta(\bx_t, t, \tau_\theta(c) \right\|^2}}. \label{eq:ldm_loss}
\end{gather}
In this process, the text description $c$ is first tokenized into textual embeddings by a Tokenizer. These textual embeddings are then passed through the CLIP text encoder $\tau_\theta(\cdot)$ to obtain text conditions. The resulting text conditions are used to guide the diffusion denoising process.

\subsection{A Comparison of CatVersion and Word Embedding Inversion}
As mentioned in Section~\ref{sec:intro}, existing inversion methods focus on learning personalized concepts in word embedding space, which can lead to the fading or even loss of concepts because the word embedding space is a feature-sparse space. Optimizing word embeddings to represent personalized concepts accurately is challenging.
Instead, we optimize the embeddings in a feature-dense space within the CLIP text encoder. We use the last three attention layers as our inversion space, as they are more feature-dense than the earlier layers due to the cumulative spatial integration in CLIP. We will demonstrate this property in Section~\ref{sec: Method}. 
Moreover, while more expressive, directly learning the token embeddings of the target concept in feature-dense space is also challenging because these embeddings tend to capture correlations with other text token embeddings. Since the Keys and Values are essential in attention blocks for learning the inter-word correlation, we concatenate the learnable embeddings to the Keys and Values, which are ultimately represented as a residual on the original attention output to learn the gap between the personalized concept and its base class. This process utilizes the existing contextual knowledge in the base class and proceeds with soft optimization from the base class to personalized concepts.
In practice, our CatVersion has demonstrated a significant enhancement in reconstruction accuracy and image editability compared to optimization in the word embedding space.

\section{Method}
\label{sec: Method}
This work uses Stable Diffusion \cite{rombach2022high} as the generative backbone. 
As shown in Figure \ref{fig:framework}, CatVersion concatenates residual embeddings within the tightly integrated feature space of the CLIP text encoder and subsequently optimizes these embeddings.
Precisely, we first locate the highly integrated feature-dense space of the text encoder to facilitate personalized concept inversion (Section~\ref{sec: clip_space}). Then, we concatenate the residual embeddings to the Keys and Values in this space and optimize them to learn the gap between the personalized concept and its base class (Section~\ref{sec: deep_embedding}). Finally, we improve the CLIP image alignment score to evaluate T2I personalization more objectively (Section~\ref{sec: metric}).

\subsection{The Feature-dense Space of CLIP}
\label{sec: clip_space}
The word embedding space is feature-sparse. The errors in word embedding learning are easily amplified when considering the correlation with free text. We aim to identify a feature-dense space in the text encoder of the diffusion model, facilitating the learning of personalized concepts.

The text condition module of stable Diffusion consists of a Tokenizer and a CLIP text encoder. The text encoder consists of 12 layers of Transformer blocks, each composed of a self-attention layer, a feed-forward layer, and normalization layers. For each text input, the Tokenizer splits it into a token sequence and then send it into the CLIP text encoder for feature encoding. The final output is used as a conditional input for the cross-attention layer of the U-Net. 
In this process, the self-attention mechanism in CLIP text encoder is crucial for learning the correlation between tokens, which is essential for understanding the semantics of the sentence. In addition, the textual features are gradually abstracted from shallow to deep layers. These claims are based on the following consensus and observation.

\noindent\textit{\textbf{Consensus: The role of the attention mechanism in Transformer.}} 
The attention mechanism \cite{vaswani2017attention} is the foundation of Transformer's powerful capabilities. It enables the model to focus on the most relevant parts of the input, allowing it to gather relevant information efficiently and accurately. The self-attention mechanism can establish connections between various parts of the input sequence, enhancing contextual understanding.

\noindent\textit{\textbf{Observation: Integration across various layers in the text encoder.}} 
We examine the integration of different self-attention layers within the text encoder of the T2I diffusion model in concept learning. More precisely, we employ the proposed CatVersion to concatenate embeddings into the self-attention layers of every two transformer blocks within the text encoder and then optimize these embeddings. We observed the same phenomenon when testing the generated results on different datasets. As depicted in Figure~\ref{fig:ablation_dog}, optimizing embeddings across all self-attention layers will overfit the target concept into the limited scene and lose its editability. Optimizing embeddings in shallow layers tends to emphasize the mastery of the simple concept, such as ``dog'', while potentially sacrificing integration with other semantic information.
When only optimizing the embeddings of intermediate modules, the results also integrate some other relatively complex concepts, such as ``Times Square''.
When optimizing embeddings in the last few modules, in addition to considering the above concepts, the result also integrates more abstract concepts, such as the action of ``running''. Therefore, as the attention layer becomes deeper, the feature integration increases. Moreover, there is a trend in information integration towards increasing abstraction, where abstraction levels increase with deeper layer positions. 

Based on the above analysis and additional experimental results, we define the feature-dense space of the CLIP text encoder as the last three self-attention layers. These layers not only play an essential role in learning the features of the target concept at different levels of abstraction but also facilitate the contextual understanding between the target concept and other semantics.

\subsection{Concatenated Embeddings Learning} 
\label{sec: deep_embedding}

Given the text feature $\mathbf{f} \in \mathbb{R}^{l \times d}$, a single-head self-attention compares the query vector $Q=W^q \mathbf{f}$ with the key vector $K=W^k \mathbf{f}$ in each Transformer block of the CLIP text encoder. Attention maps are then determined based on the similarities to indicate the importance of each input token. These attention weights are used to compute a weighted average of the value vector $V=W^v \mathbf{f}$, resulting in an output representation. This process can be expressed as follows:
\begin{gather}
\begin{aligned}
    \text{Attn}(Q, K, V) = \text{Softmax}\Big(\frac{QK^T}{\sqrt{d'}} \Big)V.
\end{aligned}\label{eq:attention_calcu}
\end{gather}
Here, $W^q$, $W^k$ and $W^v$ are projection matrix of the Query, Key and Value feature. $A=\text{Softmax} (QK^T / \sqrt{d'})$ is the attention map used to aggregate the value. 

Since we locate the feature-dense space on the last few self-attention layers of the CLIP text encoder, our goal is to implement the concept inversion in this space. Inspired by Li et al. \cite{li2021prefix}, who use prefix embeddings to indicate prompt instructions, we concatenate residual embeddings $\Delta_{k}$, $\Delta_{v} \in \mathbb{R}^{n \times d}$ to the Key and Value embeddings $K$, $V \in \mathbb{R}^{l \times d}$ for the feature-dense self-attention layers. Then, we use the new Key and Value embeddings to calculate the self-attention. This process is expressed as follows:
\begin{gather}
\begin{aligned}
    \text{Attn}(Q, K', V') = \text{Softmax}\Big(\frac{Q{K'}^T}{\sqrt{d'}} \Big){V'},
    \\\textnormal{ where } K'=W^k \mathbf{f}+\Delta_{k} \\\textnormal{ and } V'=W^v \mathbf{f}+\Delta_{v}.
\end{aligned}\label{eq:attention_calcu_new}
\end{gather}
Here, $K'$, $V' \in \mathbb{R}^{(l+n) \times d}$ are the new Key and Value embeddings which concatenated the residual embeddings. The new attantion map $A'=\text{Softmax} (QK'^T / \sqrt{d'}) \in \mathbb{R}^{l \times (l+n)}$. The attention output $\text{Attn}(Q, K', V')  \in \mathbb{R}^{l \times d}$ retains its dimensionality intact.
To optimize the residual embeddings, our overall optimization objective is derived from the simplified least squares error in Eq.~\ref{eq:ldm_loss}:
\begin{gather}
\begin{aligned}
    \Delta^{*} = \argmin \Eb{t, \bx_0, \bepsilon}{ { \left\| \bepsilon - \bepsilon_\theta(\bx_t, t, \tau_\theta(c)) \right\|^2}}, 
\end{aligned}\label{eq:total_loss}
\end{gather}
where $\Delta^{*} = \{\Delta^l_{k, v}\}_{l=i}^{j}$ represents a set of learnable residual embeddings concatenated on the Keys and Values in all feature-dense self-attention layers, from layer $i$ to $j$. 
 
During training, we use the base class of the personalized concept as text input. For instance, ``dog'' can be the base class for a specific-looking dog. Then, we concatenate residual embeddings to the Keys and Values of self-attention. When calculating the attention score, the residual embeddings are weighted averaged together with the other token embeddings, aiding the computation of correlations in the input sequence and obtaining a better feature representation. These residual embeddings ultimately manifest as a residual on the original attention output to learn the internal gap between the base class and the target concept.

Similar to word embedding inversion methods, the residual embeddings of CatVersion are plug-and-play. They can be deleted and replaced according to different personalized tasks while maintaining the integrity of the T2I generative network. Additionally, CatVersion is compatible with various improvements of word embedding inversion \cite{huang2023reversion,voynov2023p+,zhang2023prospect}, greatly accelerating its application.

\subsection{Evaluation Metric}
\label{sec: metric}
Accurate and objective quantitative evaluation is necessary for T2I personalization tasks. To achieve unbiased evaluation, both the ability to restore and edit personalized concepts need to be considered. Recently, some methods \cite{gal2022image,kumari2023multi,tewel2023key,wei2023elite,zhang2023prospect} introduce CLIP text and image alignment scores to independently assess the fidelity of generated images to free-text and the restoration of personalized concepts.
However, the CLIP image alignment score is not well-suited for evaluating the text-guided personalized results. It measures the similarity of all image features, which is susceptible to difference between the non-object parts of the reference image and the generated image.
For example, the CLIP image alignment score calculated by the method of generating images that are easily overfitted to the training scene will be very high, but the majority of the
reasons are attributed to overfitting of the background. 
To bridge this gap, We obtain the mask of the personalized concept in both the generated and reference images. Subsequently, we calculate the CLIP image alignment score for the region within the mask:
\begin{equation}
\begin{aligned}
\text{CLIP}_{\text{img+}} = \text{CLIP}_{\text{img}}(M \odot I, M_s \odot I_s).
\end{aligned}
\label{eq:evaluation_metric}
\end{equation}
Here, $M$ and $M_S$ denote the masks of the generated and the reference images, while $I$ and $I_S$ denote the generated and the reference images, respectively.
\begin{table}[t]
\centering
\resizebox{\columnwidth}{!}{
  \begin{tabular}{lccc}
    \toprule
    Method & Text Alignment $\uparrow$ & Image Alignment $\uparrow$  & Overall $\uparrow$ \\
    \midrule
    Textual Inversion \cite{gal2022image} & 0.2213 & 0.7831 & 0.4163 \\ 
    DreamBooth \cite{ruiz2023dreambooth} & 0.2227 & \textbf{0.8802} & 0.4427\\ 
    Perfusion \cite{tewel2023key} & 0.2488 & 0.7627 & 0.4356 \\
    ELITE \cite{wei2023elite} & 0.2040 & 0.8457 & 0.4154 \\
    \textbf{CatVersion} & \textbf{0.2514} & 0.8048 & \textbf{0.4498} \\ 
    \bottomrule
  \end{tabular}
}
\vspace{-0.5em}
\caption{\textbf{Quantitative Results.} We measured the average CLIP alignment scores of several methods in different personalized scenarios. Our method is significantly superior to the baseline and exhibits the highest text alignment score and balanced score.}
\label{tab:quantitative}
\end{table}

\begin{table}[t]
\centering
\resizebox{\columnwidth}{!}{
  \begin{tabular}{lccc}
    \toprule
    Method & Text Alignment $\uparrow$ & Image Alignment $\uparrow$  & Overall $\uparrow$ \\
    \midrule
    Textual Inversion \cite{gal2022image} & 10.25\% & 7.00\% & 8.25\% \\ 
    DreamBooth \cite{ruiz2023dreambooth} & 23.75\% & \textbf{36.00}\% & 24.00\% \\ 
    Perfusion \cite{tewel2023key} & 27.75\% & 20.50\% & 17.75\% \\
    ELITE \cite{wei2023elite} & 7.25\% & 13.00\% & 10.00\% \\
    \textbf{CatVersion} & \textbf{31.00\%} & 23.50\% & \textbf{40.00\%} \\ 
    \bottomrule
  \end{tabular}
}
\vspace{-0.5em}
\caption{\textbf{User Study.} We investigate the respondents' preference for the results of five different personalized generation algorithms.}
\label{tab:userstudy}
\end{table}

\section{Experiments}

\begin{figure*}[t]
  \centering
   \includegraphics[width=1\linewidth]{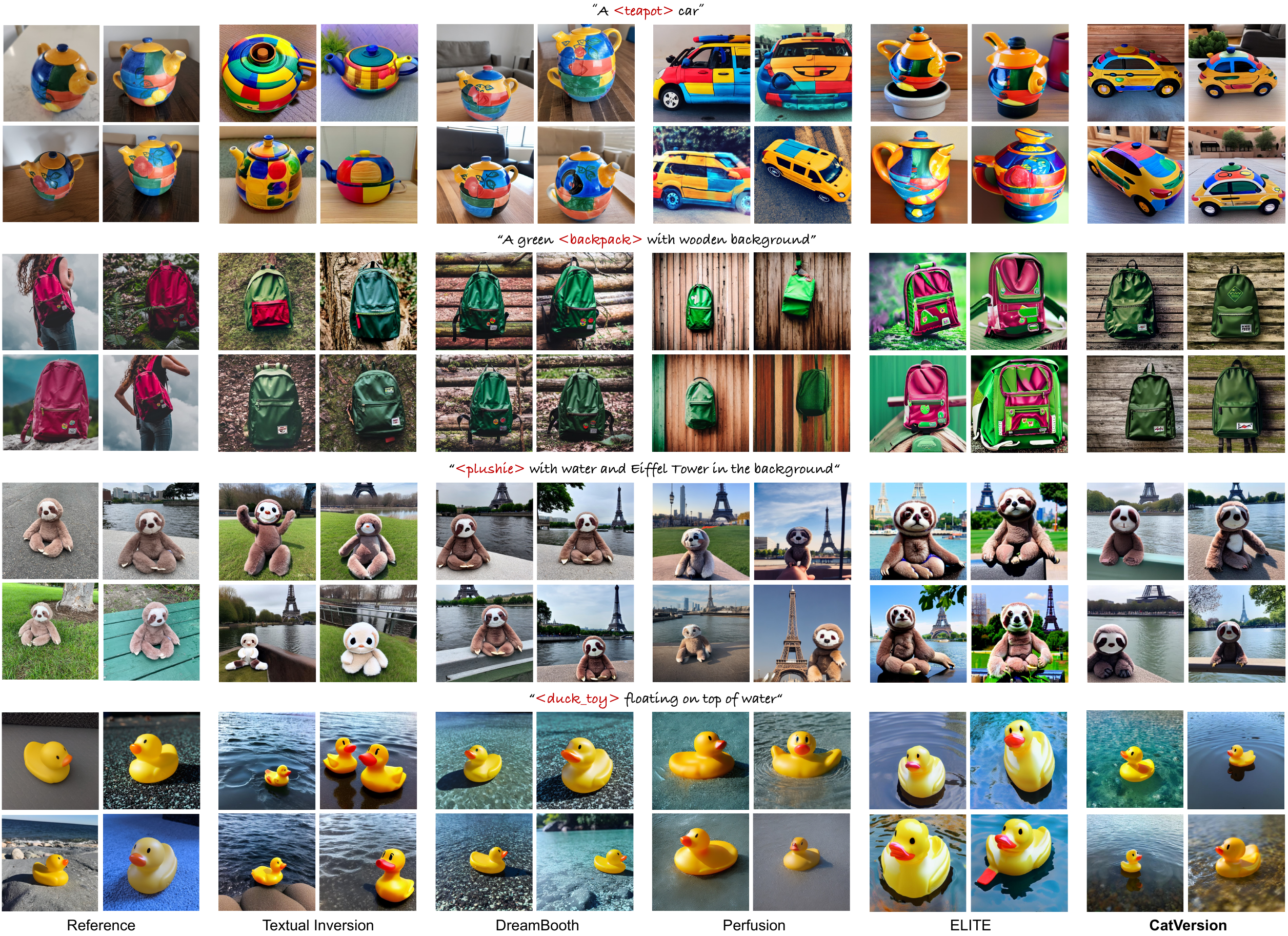}
   \vspace{-15pt}
   \caption{\textbf{Qualitative Comparisons with Existing Methods}. Our CatVersion more faithfully restores personalized concepts and achieves more powerful editing capabilities in the combination of various concepts and free text.
   }
   \label{fig:comparison1}
\end{figure*}

\begin{figure}[t]
  \centering
   \includegraphics[width=1\linewidth]{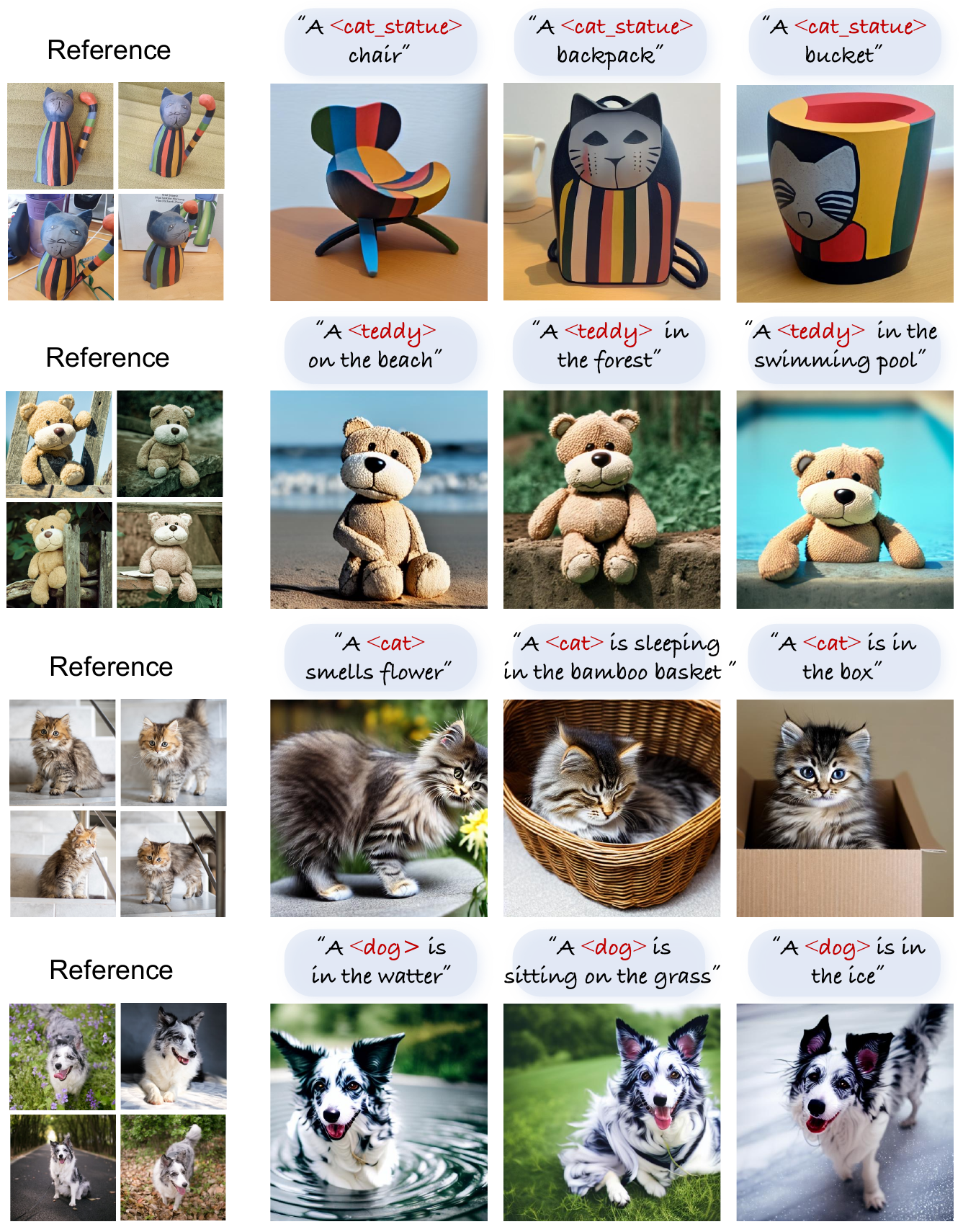}
   \vspace{-20pt}
   \caption{\textbf{Visualization Results of Our Method}. Our CatVersion achieves a better balance between faithful reconstruction of the target concept and more robust editability.
   }
   \label{fig:ours_result}
\end{figure}

\begin{figure}[t]
  \centering
   \includegraphics[width=1\linewidth]{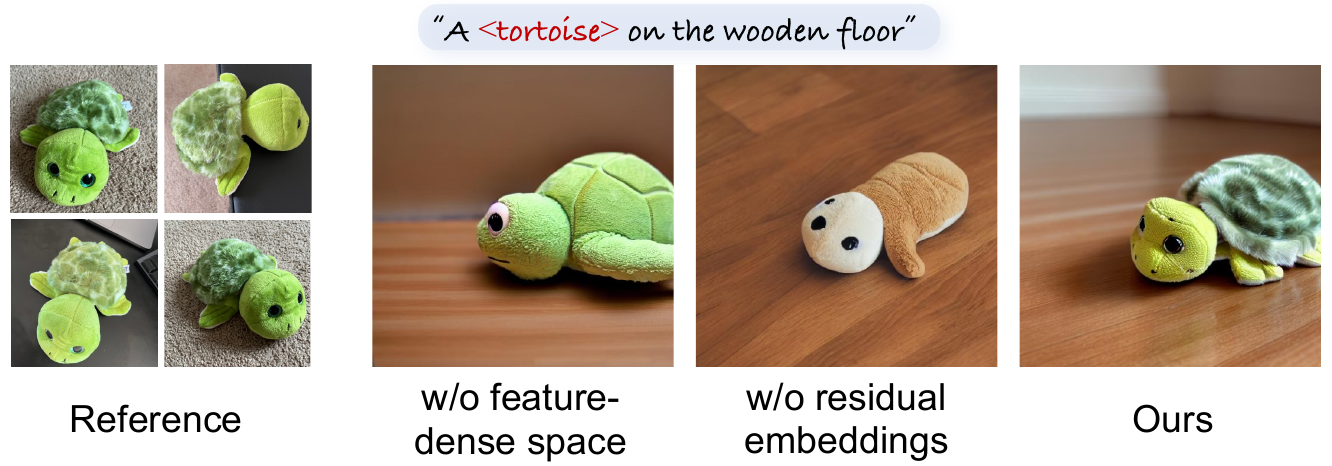}
   \vspace{-20pt}
   \caption{\textbf{Visualization of Ablation Study}. We visualize the impact of removing our two configurations separately on the results, intuitively showcasing the importance of these configurations.
   }
   \label{fig:ablation}
\end{figure}

In this section, we present qualitative and quantitative results. We demonstrate that our proposed CatVersion outperforms baseline Textual Inversion and is highly competitive compared to state-of-the-art methods.

\noindent \textbf{Compared Methods}.
We compare our CatVersion with the state-of-the-art competitors Textual Inversion \cite{gal2022image}, DreamBooth \cite{ruiz2023dreambooth}, Perfusion \cite{tewel2023key}, and ELITE \cite{wei2023elite}. 
We use Stable Diffusion v1.5 \cite{rombach2022high} as the generative backbone, and the generated image resolution is $512\times512$. For Textual Inversion and DreamBooth, we employ their Diffusers version \cite{von-platen-etal-2022-diffusers}, while for perfusion and ELITE, we utilize their official implementation. All experiments are conducted following the official recommended configurations.

\noindent \textbf{Implementation Details}.
The implementation is based on one NVIDIA GeForce RTX 3090 GPU. We employed a pre-trained Stable Diffusion and concatenated residual embeddings $\Delta_{k}$, $\Delta_{v} \in \mathbb{R}^{n \times d}$ to the Keys and Values within the last three self-attention layers of the CLIP text encoder, with $n=1$ and $d=768$. We optimized these embeddings employing a learning rate of 0.005 and a batch size of 1 for around 1000 steps. During the inference process, we employed the DDIM sampler with the sampling steps $T=50$ and the classifier free guidance with guidance scale $w=10$. 

\subsection{Qualitative Comparisons}
Figure~\ref{fig:comparison1} shows the generated results of CatVersion and its competitors by applying personalized concepts to different scenarios through free text descriptions. CatVersion significantly outperforms the baseline Textual inversion in terms of restoring the personalized concept and the text-guided editability. At the same time, Textual inversion and Perfusion suffer from concept degradation while struggling to incorporate the context. DreamBooth and ELITE tend to overfit to the training set and exhibit limited editability. In contrast, CatVersion exhibits a more balanced performance in terms of reconstructing personalized concepts and enabling free text-guided editing capabilities.

\begin{table}[t]
\centering
\resizebox{\columnwidth}{!}{
  \begin{tabular}{lccc}
    \toprule
    Method & Text Alignment$ \uparrow$  & Image Alignment$ \uparrow$  & Overall $\uparrow$ \\ 
    \midrule
    w/o feature-dense space  & 0.2386 & 0.7925 & 0.4348 \\ 
    w/o residual embeddings & \textbf{0.2651} & 0.7327 & 0.4407 \\ 
    \textbf{Ours} & 0.2514 & \textbf{0.8048} & \textbf{0.4498} \\ 
    \bottomrule
  \end{tabular}
}
\vspace{-0.5em}
\caption{\textbf{Ablation Study}. We validate the independent impact of our proposed feature-dense space and residual embeddings on the results, emphasizing the importance of these two configurations.}
\label{tab:ablation}
\end{table}

\subsection{Quantitative Comparisons}
Quantitative analysis is performed based on paired CLIP alignment scores. The CLIP text alignment score calculates the similarity between the text and the generated image to evaluate the editability of personalized generation. We adjust the CLIP image alignment score to focus on the similarity of the personalized concepts between the generated and the reference image to evaluate the reconstruction fidelity. Additionally, we employ the geometric mean of two scores to evaluate the overall generation ability. We compare the average paired CLIP alignment scores between the results of CatVersion and those of its competitors on datasets selected from these competitors \cite{gal2022image,ruiz2023dreambooth,tewel2023key}. The guided text includes four editing dimensions, which are attribute transfer, scene transformation, action editing, and concept addition, for achieving unbiased evaluation.
As shown in Table~\ref{tab:quantitative},
Our CatVersion outperforms the baseline Textual Inversion in both fidelity and editability.
Additionally, DreamBooth and ELITE are prone to overfitting to the training images, resulting in high reconstruction scores at the massive expense of editability scores. Although Perfusion demonstrates superiority in editability, it delivers unsatisfactory results in terms of reconstruction fidelity. In comparison, our CatVersion outperforms these competitors in both editability and overall personalized effect.

\noindent \textbf{User Study.}
We assess the mean preference of 50 participants for CatVersion and four other state-of-the-art methods. Each participant is required to complete a survey consisting of 8 random questions. Table~\ref{tab:userstudy} indicates user prefer CatVersion in text-guided editability and overall personalized effect, with CatVersion ranking second only to DreamBooth in reconstruction fidelity.


\subsection{Ablation Study}
 We conducted ablation studies to validate the effectiveness of our proposed feature-dense space inversion and residual embeddings learning. As depicted in Table~\ref{tab:ablation}, the feature-dense space inversion notably enhances paired CLIP alignment scores. This suggests that optimizing the target concept in feature-dense spaces is more effective for learning the concept itself and improving contextual understanding. Additionally, concatenating residual embeddings to the keys and values of the CLIP text encoder resulted in a significant enhancement of the CLIP image alignment score, indicating a substantial improvement in the reconstruction ability of the target concept, as shown in Figure~\ref{fig:ablation}. 

\subsection{Limitations}
CatVersion still needs to be optimized separately for each concept, so its inversion speed is not as fast as encoder based methods. In addition, CatVersion can only learn one concept in a single optimization process, thereby limiting its applicability in specific tasks.

\section{Conclusion}
In this paper, we propose CatVersion to achieve accurate and efficient personalized text-to-image generation. Differing from existing methods, CatVersion learns the gap between the personalized concept and its base class by concatenating residual embeddings to the Keys and Values of the feature-dense layers in the CLIP text encoder, restoring personalized concepts more faithfully and enables more robust editing. 
We also dissect the integration of the T2I diffusion model's text encoder to locate feature-dense inversion spaces. We believe that our work provides new insights for concept inversion. We hope that it will inspire future research in terms of inversion-based generation and editing.

{
    \small
    \bibliographystyle{ieeenat_fullname}
    \bibliography{main}

\begin{thebibliography}{38}
\providecommand{\natexlab}[1]{#1}
\providecommand{\url}[1]{\texttt{#1}}
\expandafter\ifx\csname urlstyle\endcsname\relax
  \providecommand{\doi}[1]{doi: #1}\else
  \providecommand{\doi}{doi: \begingroup \urlstyle{rm}\Url}\fi

\bibitem[Dhariwal and Nichol(2021)]{dhariwal2021diffusion}
Prafulla Dhariwal and Alexander Nichol.
\newblock Diffusion models beat gans on image synthesis.
\newblock \emph{Advances in neural information processing systems}, 34:\penalty0 8780--8794, 2021.

\bibitem[Ding et~al.(2021)Ding, Yang, Hong, Zheng, Zhou, Yin, Lin, Zou, Shao, Yang, et~al.]{ding2021cogview}
Ming Ding, Zhuoyi Yang, Wenyi Hong, Wendi Zheng, Chang Zhou, Da Yin, Junyang Lin, Xu Zou, Zhou Shao, Hongxia Yang, et~al.
\newblock Cogview: Mastering text-to-image generation via transformers.
\newblock \emph{Advances in Neural Information Processing Systems}, 34:\penalty0 19822--19835, 2021.

\bibitem[Gafni et~al.(2022)Gafni, Polyak, Ashual, Sheynin, Parikh, and Taigman]{gafni2022make}
Oran Gafni, Adam Polyak, Oron Ashual, Shelly Sheynin, Devi Parikh, and Yaniv Taigman.
\newblock Make-a-scene: Scene-based text-to-image generation with human priors.
\newblock In \emph{European Conference on Computer Vision}, pages 89--106. Springer, 2022.

\bibitem[Gal et~al.(2022)Gal, Alaluf, Atzmon, Patashnik, Bermano, Chechik, and Cohen-Or]{gal2022image}
Rinon Gal, Yuval Alaluf, Yuval Atzmon, Or Patashnik, Amit~H Bermano, Gal Chechik, and Daniel Cohen-Or.
\newblock An image is worth one word: Personalizing text-to-image generation using textual inversion.
\newblock \emph{arXiv preprint arXiv:2208.01618}, 2022.

\bibitem[Gal et~al.(2023)Gal, Arar, Atzmon, Bermano, Chechik, and Cohen-Or]{gal2023encoder}
Rinon Gal, Moab Arar, Yuval Atzmon, Amit~H Bermano, Gal Chechik, and Daniel Cohen-Or.
\newblock Encoder-based domain tuning for fast personalization of text-to-image models.
\newblock \emph{ACM Transactions on Graphics (TOG)}, 42\penalty0 (4):\penalty0 1--13, 2023.

\bibitem[Han et~al.(2023)Han, Li, Zhang, Milanfar, Metaxas, and Yang]{han2023svdiff}
Ligong Han, Yinxiao Li, Han Zhang, Peyman Milanfar, Dimitris Metaxas, and Feng Yang.
\newblock Svdiff: Compact parameter space for diffusion fine-tuning.
\newblock \emph{arXiv preprint arXiv:2303.11305}, 2023.

\bibitem[Ho et~al.(2020)Ho, Jain, and Abbeel]{ho2020denoising}
Jonathan Ho, Ajay Jain, and Pieter Abbeel.
\newblock Denoising diffusion probabilistic models.
\newblock \emph{Advances in neural information processing systems}, 33:\penalty0 6840--6851, 2020.

\bibitem[Huang et~al.(2023)Huang, Wu, Jiang, Chan, and Liu]{huang2023reversion}
Ziqi Huang, Tianxing Wu, Yuming Jiang, Kelvin~CK Chan, and Ziwei Liu.
\newblock Reversion: Diffusion-based relation inversion from images.
\newblock \emph{arXiv preprint arXiv:2303.13495}, 2023.

\bibitem[Kumari et~al.(2023)Kumari, Zhang, Zhang, Shechtman, and Zhu]{kumari2023multi}
Nupur Kumari, Bingliang Zhang, Richard Zhang, Eli Shechtman, and Jun-Yan Zhu.
\newblock Multi-concept customization of text-to-image diffusion.
\newblock In \emph{Proceedings of the IEEE/CVF Conference on Computer Vision and Pattern Recognition}, pages 1931--1941, 2023.

\bibitem[Li and Liang(2021)]{li2021prefix}
Xiang~Lisa Li and Percy Liang.
\newblock Prefix-tuning: Optimizing continuous prompts for generation.
\newblock In \emph{Proceedings of the 59th Annual Meeting of the Association for Computational Linguistics and the 11th International Joint Conference on Natural Language Processing (Volume 1: Long Papers)}, pages 4582--4597, 2021.

\bibitem[Lin et~al.(2021)Lin, Men, Yang, Zhou, Ding, Zhang, Wang, Wang, Jiang, Jia, et~al.]{lin2021m6}
Junyang Lin, Rui Men, An Yang, Chang Zhou, Ming Ding, Yichang Zhang, Peng Wang, Ang Wang, Le Jiang, Xianyan Jia, et~al.
\newblock M6: A chinese multimodal pretrainer.
\newblock \emph{arXiv preprint arXiv:2103.00823}, 2021.

\bibitem[Meng et~al.(2022)Meng, Bau, Andonian, and Belinkov]{meng2022locating}
Kevin Meng, David Bau, Alex Andonian, and Yonatan Belinkov.
\newblock Locating and editing factual associations in gpt.
\newblock \emph{Advances in Neural Information Processing Systems}, 35:\penalty0 17359--17372, 2022.

\bibitem[Nichol et~al.(2021)Nichol, Dhariwal, Ramesh, Shyam, Mishkin, McGrew, Sutskever, and Chen]{nichol2021glide}
Alex Nichol, Prafulla Dhariwal, Aditya Ramesh, Pranav Shyam, Pamela Mishkin, Bob McGrew, Ilya Sutskever, and Mark Chen.
\newblock Glide: Towards photorealistic image generation and editing with text-guided diffusion models.
\newblock \emph{arXiv preprint arXiv:2112.10741}, 2021.

\bibitem[Nichol and Dhariwal(2021)]{nichol2021improved}
Alexander~Quinn Nichol and Prafulla Dhariwal.
\newblock Improved denoising diffusion probabilistic models.
\newblock In \emph{International Conference on Machine Learning}, pages 8162--8171. PMLR, 2021.

\bibitem[Qiao et~al.(2019)Qiao, Zhang, Xu, and Tao]{qiao2019mirrorgan}
Tingting Qiao, Jing Zhang, Duanqing Xu, and Dacheng Tao.
\newblock Mirrorgan: Learning text-to-image generation by redescription.
\newblock In \emph{Proceedings of the IEEE/CVF Conference on Computer Vision and Pattern Recognition}, pages 1505--1514, 2019.

\bibitem[Radford et~al.(2021)Radford, Kim, Hallacy, Ramesh, Goh, Agarwal, Sastry, Askell, Mishkin, Clark, et~al.]{radford2021learning}
Alec Radford, Jong~Wook Kim, Chris Hallacy, Aditya Ramesh, Gabriel Goh, Sandhini Agarwal, Girish Sastry, Amanda Askell, Pamela Mishkin, Jack Clark, et~al.
\newblock Learning transferable visual models from natural language supervision.
\newblock In \emph{International conference on machine learning}, pages 8748--8763. PMLR, 2021.

\bibitem[Ramesh et~al.(2021)Ramesh, Pavlov, Goh, Gray, Voss, Radford, Chen, and Sutskever]{ramesh2021zero}
Aditya Ramesh, Mikhail Pavlov, Gabriel Goh, Scott Gray, Chelsea Voss, Alec Radford, Mark Chen, and Ilya Sutskever.
\newblock Zero-shot text-to-image generation.
\newblock In \emph{International Conference on Machine Learning}, pages 8821--8831. PMLR, 2021.

\bibitem[Ramesh et~al.(2022)Ramesh, Dhariwal, Nichol, Chu, and Chen]{ramesh2022hierarchical}
Aditya Ramesh, Prafulla Dhariwal, Alex Nichol, Casey Chu, and Mark Chen.
\newblock Hierarchical text-conditional image generation with clip latents.
\newblock \emph{arXiv preprint arXiv:2204.06125}, 1\penalty0 (2):\penalty0 3, 2022.

\bibitem[Rombach et~al.(2022)Rombach, Blattmann, Lorenz, Esser, and Ommer]{rombach2022high}
Robin Rombach, Andreas Blattmann, Dominik Lorenz, Patrick Esser, and Bj{\"o}rn Ommer.
\newblock High-resolution image synthesis with latent diffusion models.
\newblock In \emph{Proceedings of the IEEE/CVF conference on computer vision and pattern recognition}, pages 10684--10695, 2022.

\bibitem[Ruiz et~al.(2023)Ruiz, Li, Jampani, Pritch, Rubinstein, and Aberman]{ruiz2023dreambooth}
Nataniel Ruiz, Yuanzhen Li, Varun Jampani, Yael Pritch, Michael Rubinstein, and Kfir Aberman.
\newblock Dreambooth: Fine tuning text-to-image diffusion models for subject-driven generation.
\newblock In \emph{Proceedings of the IEEE/CVF Conference on Computer Vision and Pattern Recognition}, pages 22500--22510, 2023.

\bibitem[Saharia et~al.(2022{\natexlab{a}})Saharia, Chan, Chang, Lee, Ho, Salimans, Fleet, and Norouzi]{saharia2022palette}
Chitwan Saharia, William Chan, Huiwen Chang, Chris Lee, Jonathan Ho, Tim Salimans, David Fleet, and Mohammad Norouzi.
\newblock Palette: Image-to-image diffusion models.
\newblock In \emph{ACM SIGGRAPH 2022 Conference Proceedings}, pages 1--10, 2022{\natexlab{a}}.

\bibitem[Saharia et~al.(2022{\natexlab{b}})Saharia, Chan, Saxena, Li, Whang, Denton, Ghasemipour, Gontijo~Lopes, Karagol~Ayan, Salimans, et~al.]{saharia2022photorealistic}
Chitwan Saharia, William Chan, Saurabh Saxena, Lala Li, Jay Whang, Emily~L Denton, Kamyar Ghasemipour, Raphael Gontijo~Lopes, Burcu Karagol~Ayan, Tim Salimans, et~al.
\newblock Photorealistic text-to-image diffusion models with deep language understanding.
\newblock \emph{Advances in Neural Information Processing Systems}, 35:\penalty0 36479--36494, 2022{\natexlab{b}}.

\bibitem[Song et~al.(2020{\natexlab{a}})Song, Meng, and Ermon]{song2020denoising}
Jiaming Song, Chenlin Meng, and Stefano Ermon.
\newblock Denoising diffusion implicit models.
\newblock \emph{arXiv preprint arXiv:2010.02502}, 2020{\natexlab{a}}.

\bibitem[Song et~al.(2020{\natexlab{b}})Song, Sohl-Dickstein, Kingma, Kumar, Ermon, and Poole]{song2020score}
Yang Song, Jascha Sohl-Dickstein, Diederik~P Kingma, Abhishek Kumar, Stefano Ermon, and Ben Poole.
\newblock Score-based generative modeling through stochastic differential equations.
\newblock \emph{arXiv preprint arXiv:2011.13456}, 2020{\natexlab{b}}.

\bibitem[Tao et~al.(2022)Tao, Tang, Wu, Jing, Bao, and Xu]{tao2022df}
Ming Tao, Hao Tang, Fei Wu, Xiao-Yuan Jing, Bing-Kun Bao, and Changsheng Xu.
\newblock Df-gan: A simple and effective baseline for text-to-image synthesis.
\newblock In \emph{Proceedings of the IEEE/CVF Conference on Computer Vision and Pattern Recognition}, pages 16515--16525, 2022.

\bibitem[Tewel et~al.(2023)Tewel, Gal, Chechik, and Atzmon]{tewel2023key}
Yoad Tewel, Rinon Gal, Gal Chechik, and Yuval Atzmon.
\newblock Key-locked rank one editing for text-to-image personalization.
\newblock In \emph{ACM SIGGRAPH 2023 Conference Proceedings}, pages 1--11, 2023.

\bibitem[Van Den~Oord et~al.(2017)Van Den~Oord, Vinyals, et~al.]{van2017neural}
Aaron Van Den~Oord, Oriol Vinyals, et~al.
\newblock Neural discrete representation learning.
\newblock \emph{Advances in neural information processing systems}, 30, 2017.

\bibitem[Vaswani et~al.(2017)Vaswani, Shazeer, Parmar, Uszkoreit, Jones, Gomez, Kaiser, and Polosukhin]{vaswani2017attention}
Ashish Vaswani, Noam Shazeer, Niki Parmar, Jakob Uszkoreit, Llion Jones, Aidan~N Gomez, {\L}ukasz Kaiser, and Illia Polosukhin.
\newblock Attention is all you need.
\newblock \emph{Advances in neural information processing systems}, 30, 2017.

\bibitem[von Platen et~al.(2022)von Platen, Patil, Lozhkov, Cuenca, Lambert, Rasul, Davaadorj, and Wolf]{von-platen-etal-2022-diffusers}
Patrick von Platen, Suraj Patil, Anton Lozhkov, Pedro Cuenca, Nathan Lambert, Kashif Rasul, Mishig Davaadorj, and Thomas Wolf.
\newblock Diffusers: State-of-the-art diffusion models.
\newblock \url{https://github.com/huggingface/diffusers}, 2022.

\bibitem[Voynov et~al.(2023)Voynov, Chu, Cohen-Or, and Aberman]{voynov2023p+}
Andrey Voynov, Qinghao Chu, Daniel Cohen-Or, and Kfir Aberman.
\newblock $ p+ $: Extended textual conditioning in text-to-image generation.
\newblock \emph{arXiv preprint arXiv:2303.09522}, 2023.

\bibitem[Wei et~al.(2023)Wei, Zhang, Ji, Bai, Zhang, and Zuo]{wei2023elite}
Yuxiang Wei, Yabo Zhang, Zhilong Ji, Jinfeng Bai, Lei Zhang, and Wangmeng Zuo.
\newblock Elite: Encoding visual concepts into textual embeddings for customized text-to-image generation.
\newblock \emph{arXiv preprint arXiv:2302.13848}, 2023.

\bibitem[Wen et~al.(2023)Wen, Jain, Kirchenbauer, Goldblum, Geiping, and Goldstein]{wen2023hard}
Yuxin Wen, Neel Jain, John Kirchenbauer, Micah Goldblum, Jonas Geiping, and Tom Goldstein.
\newblock Hard prompts made easy: Gradient-based discrete optimization for prompt tuning and discovery.
\newblock \emph{arXiv preprint arXiv:2302.03668}, 2023.

\bibitem[Xu et~al.(2018)Xu, Zhang, Huang, Zhang, Gan, Huang, and He]{xu2018attngan}
Tao Xu, Pengchuan Zhang, Qiuyuan Huang, Han Zhang, Zhe Gan, Xiaolei Huang, and Xiaodong He.
\newblock {AttnGAN}: Fine-grained text to image generation with attentional generative adversarial networks.
\newblock In \emph{CVPR}, 2018.

\bibitem[Yin et~al.(2019)Yin, Liu, Sheng, Yu, Wang, and Shao]{yin2019semantics}
Guojun Yin, Bin Liu, Lu Sheng, Nenghai Yu, Xiaogang Wang, and Jing Shao.
\newblock Semantics disentangling for text-to-image generation.
\newblock In \emph{Proceedings of the IEEE/CVF conference on computer vision and pattern recognition}, pages 2327--2336, 2019.

\bibitem[Zhang et~al.(2017)Zhang, Xu, Li, Zhang, Wang, Huang, and Metaxas]{zhang2017stackgan}
Han Zhang, Tao Xu, Hongsheng Li, Shaoting Zhang, Xiaogang Wang, Xiaolei Huang, and Dimitris~N Metaxas.
\newblock Stackgan: Text to photo-realistic image synthesis with stacked generative adversarial networks.
\newblock In \emph{Proceedings of the IEEE international conference on computer vision}, pages 5907--5915, 2017.

\bibitem[Zhang et~al.(2018)Zhang, Xu, Li, Zhang, Wang, Huang, and Metaxas]{zhang2018stackgan++}
Han Zhang, Tao Xu, Hongsheng Li, Shaoting Zhang, Xiaogang Wang, Xiaolei Huang, and Dimitris~N Metaxas.
\newblock Stackgan++: Realistic image synthesis with stacked generative adversarial networks.
\newblock \emph{IEEE transactions on pattern analysis and machine intelligence}, 41\penalty0 (8):\penalty0 1947--1962, 2018.

\bibitem[Zhang et~al.(2023)Zhang, Dong, Tang, Huang, Huang, Ma, Lee, Deussen, and Xu]{zhang2023prospect}
Yuxin Zhang, Weiming Dong, Fan Tang, Nisha Huang, Haibin Huang, Chongyang Ma, Tong-Yee Lee, Oliver Deussen, and Changsheng Xu.
\newblock Prospect: Expanded conditioning for the personalization of attribute-aware image generation.
\newblock \emph{arXiv preprint arXiv:2305.16225}, 2023.

\bibitem[Zhu et~al.(2019)Zhu, Pan, Chen, and Yang]{zhu2019dm}
Minfeng Zhu, Pingbo Pan, Wei Chen, and Yi Yang.
\newblock Dm-gan: Dynamic memory generative adversarial networks for text-to-image synthesis.
\newblock In \emph{Proceedings of the IEEE/CVF conference on computer vision and pattern recognition}, pages 5802--5810, 2019.

\end{thebibliography}
}







\end{document}